%% file: main.tex
\documentclass[conference]{IEEEtran}

\input{preamble}

\begin{document}

\title{\framework: Aligned Latent Embedding Retrieval for Time Series Imputation}

\author{
\IEEEauthorblockN{Xuan-Thong Truong\textsuperscript{1}, Trung-Kien Le\textsuperscript{1}, Tung Kieu\textsuperscript{2}, Thi-Thu Nguyen\textsuperscript{1}, Nhat-Hai Nguyen\textsuperscript{1,*}}

\IEEEauthorblockA{\textsuperscript{1}\textit{School of Computer Science, Hanoi University of Science and Technology, Hanoi, Vietnam}}
\IEEEauthorblockA{\textsuperscript{2}\textit{Department of Computer Science, Aalborg University, Aalborg, Denmark}}
\IEEEauthorblockA{\textsuperscript{*}Corresponding author: hai.nguyennhat@hust.edu.vn} 
}

\maketitle

\input{revised_sections/abstract}
\input{revised_sections/introduction}
\input{revised_sections/related_work}
\input{revised_sections/methodology}
\input{revised_sections/experiments}
\input{revised_sections/conlusion}

\bibliographystyle{IEEEtranS}
\bibliography{revised_references}

\end{document}

%% file: preamble.tex
\IEEEoverridecommandlockouts

\usepackage[table,xcdraw]{xcolor}
\usepackage{cite}
\usepackage{amsmath,amssymb,amsfonts}
\usepackage{algorithmic}
\usepackage{graphicx}
\usepackage{textcomp}
\usepackage{multirow}
\usepackage{microtype}
\usepackage{subcaption}
\usepackage[normalem]{ulem}
\usepackage[table,xcdraw]{xcolor}
\usepackage{tabularx}
\usepackage{booktabs}
\usepackage[font=small]{caption}
\usepackage{array}
\usepackage{xspace}
\usepackage{dblfloatfix}

\def\tablename{Table}

\newcommand{\framework}{\texttt{ALER-TI}\xspace}

\newcolumntype{C}{>{\centering\arraybackslash}X}

\usepackage[colorlinks=true, allcolors=blue]{hyperref}

\def\BibTeX{{\rm B\kern-.05em{\sc i\kern-.025em b}\kern-.08em
    T\kern-.1667em\lower.7ex\hbox{E}\kern-.125emX}}

%% file: revised_sections/abstract.tex
\begin{abstract}
Deep learning has significantly advanced time series imputation, yet most existing architectures primarily rely on localized temporal context within the corrupted input sequence. This reliance can be limiting in real-world scenarios, where time series often exhibit non-stationary dynamics, weak temporal correlations, and infrequent patterns that are difficult to reconstruct from nearby observations alone. In this paper, we propose \framework, Aligned Latent Embedding Retrieval for Time Series Imputation, a retrieval-augmented framework that explicitly leverages historical patterns to supplement degraded local context for more reliable missing-value reconstruction. The core of \framework is Latent Embedding Alignment (LEA), which mitigates the representation mismatch between corrupted queries and complete historical candidates. By applying post-hoc masking in the latent space, LEA aligns candidates with the query's missingness pattern while allowing historical embeddings to be pre-computed and cached for efficient retrieval. \framework is model-agnostic and can be integrated with various imputation backbones through a lightweight adaptation module. Extensive experiments on six real-world datasets under different missing rates demonstrate that \framework consistently improves strong baseline models and enhances robustness across diverse imputation settings.
\end{abstract}


%% file: revised_sections/introduction.tex
\section{Introduction}
\label{sec:introduction}

Time series data collected from real-world sensors and monitoring systems are frequently affected by missing values due to sensor malfunctions, transmission failures, or human errors~\cite{gong2021missing}. Since many downstream analytical models require complete data matrices as inputs, missing observations can severely undermine the reliability of critical applications, ranging from healthcare monitoring~\cite{che2018recurrent, nguyen2026spectral} and financial forecasting~\cite{abu1996introduction} to industrial anomaly detection~\cite{canizo2019multi}. To address this challenge, various deep learning architectures have been developed for time series imputation, including \texttt{CNN}-based models~\cite{wu2022timesnet, luo2024moderntcn}, \texttt{Transformer} variants~\cite{du2023saits, liu2022non}, and decomposition-based networks~\cite{zeng2023transformers, wu2021autoformer}. Despite their different architectural designs, these methods mainly rely on the localized temporal context within the corrupted input sequence to reconstruct missing observations.

\begin{figure}[t]
\centering
\includegraphics[width=1.0\linewidth]{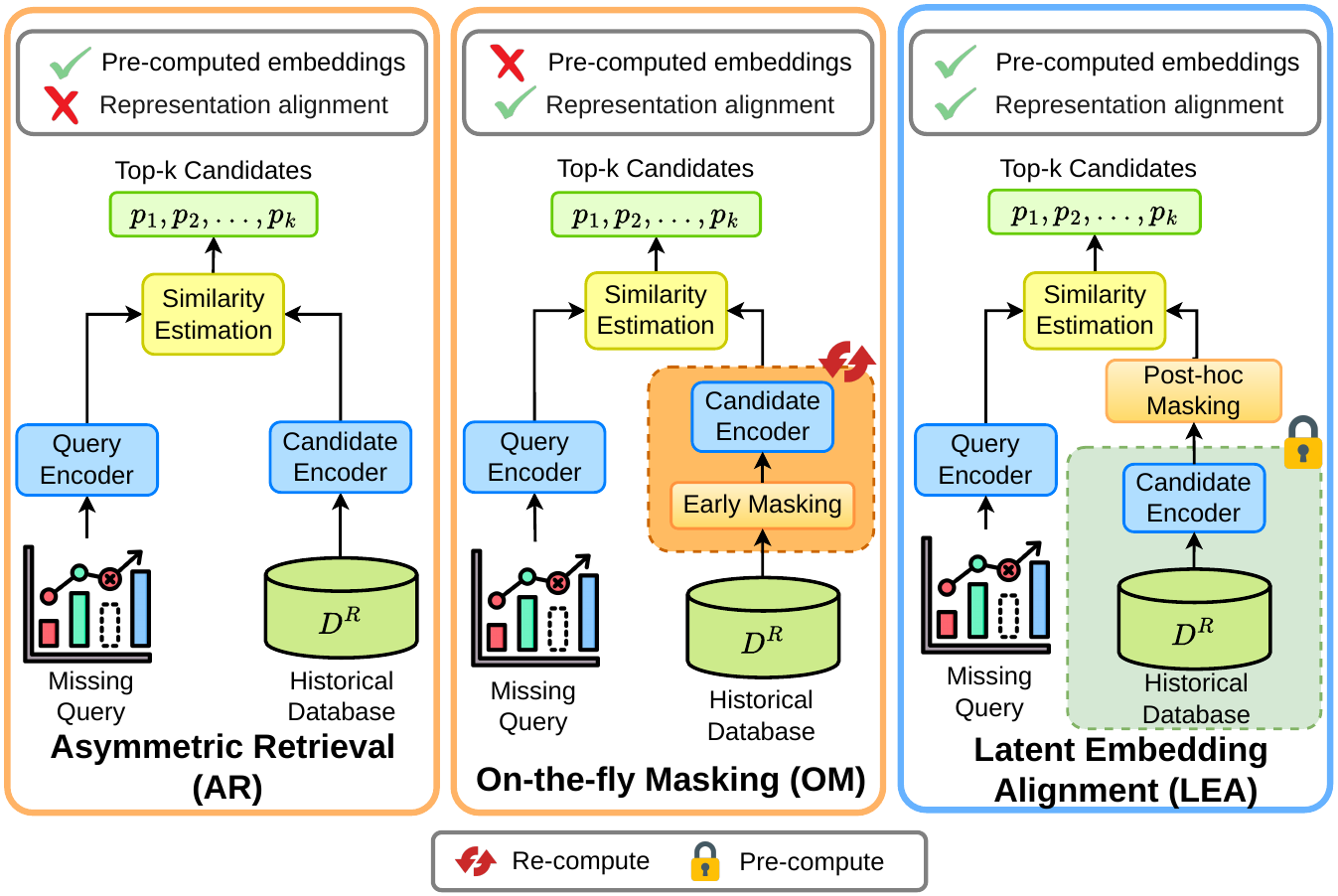}
\caption{\textbf{Comparison of retrieval strategies for time-series imputation.}
\textbf{(a) Asymmetric Retrieval (AR):} Encodes the query and candidates independently, with candidates represented as fully observed trajectories.
\textbf{(b) On-the-fly Masking (OM):} Applies masking before encoding to align candidate representations with the query pattern, but requires online re-encoding at inference time.
\textbf{(c) LEA (Ours):} Uses mask-agnostic encoding and performs masking in latent space, enabling reuse of pre-computed embeddings while preserving query-aware alignment.}
\label{fig:intro_image}
\end{figure}

However, real-world time series often exhibit complex and non-stationary dynamics driven by non-deterministic processes, where temporal correlations may weaken over time~\cite{liu2022non, wu2026out}. Such dynamics can lead to infrequent patterns and distributional variations, making it difficult for models to infer missing values from the observed local context alone. This limitation is especially problematic when the corrupted segment corresponds to a rare event or when nearby observations provide insufficient correlated information for reliable reconstruction.

A promising direction is to leverage historical patterns through retrieval-augmented mechanisms, a paradigm that has recently shown strong potential in time series forecasting~\cite{nguyen2026vardiff, liu2024retrieval, du2026predicting}. For imputation, retrieval is especially appealing because it provides explicit reference patterns at inference time, complementing the information stored implicitly in model parameters. When the local context around missing values is weak, ambiguous, or affected by non-stationary dynamics, similar historical segments may offer useful structural evidence for reconstruction. Retrieval can also help reuse rare but relevant temporal patterns when similar situations reappear, which is difficult for fixed-parameter models to achieve when such patterns occur only sparsely in the training data~\cite{arpit2017closer, zhou2021informer}.

However, retrieval for time series imputation is fundamentally different from retrieval for forecasting. In forecasting, the query sequence is usually fully observed, so similarity can be estimated from a complete input context. In imputation, by contrast, the query itself is corrupted. Missing values directly distort the query representation and make it difficult to compare the query with complete historical candidates. Therefore, the key challenge is not merely how to retrieve historical patterns, but how to measure similarity between a corrupted query and clean candidates in a representation space that is both aligned and efficient.

As illustrated in Fig.~\ref{fig:intro_image}(a), a naive Asymmetric Retrieval strategy encodes the corrupted query and complete candidates independently. This design allows candidate embeddings to be pre-computed, but it creates a representation mismatch: the query encoder is affected by missingness artifacts, whereas the candidate encoder processes fully observed trajectories. Consequently, the resulting similarity scores may not reflect the true structural relationship between signals, causing the retriever to select irrelevant historical patterns. A more aligned alternative is On-the-fly Masking, shown in Fig.~\ref{fig:intro_image}(b), which applies the query-specific mask to each candidate before encoding. Although this strategy reduces the representation gap, it destroys the efficiency of cached retrieval because candidate representations become query-dependent and must be recomputed online.

To address these challenges, we propose \framework, a retrieval-augmented framework for time series imputation centered on \textit{Latent Embedding Alignment} (LEA). As illustrated in Fig.~\ref{fig:intro_image}(c), LEA adopts a mask-agnostic encoding paradigm that shifts the interaction with missingness patterns from raw input encoding to a downstream latent alignment stage. This design allows historical candidates to be encoded and cached offline, while LEA applies post-hoc masking and context-aware interaction to align corrupted queries with the pre-indexed candidate space during retrieval. In this way, \framework preserves the efficiency of cached retrieval while mitigating the representation mismatch between corrupted queries and clean historical references.

We further design \framework as a model-agnostic framework that can be integrated with various time series imputation backbones, such as \texttt{ModernTCN}~\cite{luo2024moderntcn}, \texttt{TimesNet}~\cite{wu2022timesnet}, and \texttt{SAITS}~\cite{du2023saits}. By augmenting local-context models with relevant historical patterns, \framework provides additional global evidence for reconstruction and improves imputation performance across different sequence lengths and missing rates.

Our primary contributions are summarized as follows:
\begin{itemize}
\item We propose \textit{Latent Embedding Alignment} (LEA), a retrieval mechanism that mitigates the representation mismatch between corrupted queries and clean historical candidates while avoiding the computational bottleneck of online database re-encoding.

\item We introduce \framework, a model-agnostic retrieval-augmented framework for time series imputation that integrates relevant historical patterns with existing backbone models through a lightweight adaptation module.

\item Extensive experiments on multiple real-world benchmarks demonstrate that \framework consistently improves strong imputation backbones and remains robust across diverse sequence lengths and missing rates.

\end{itemize}

%% file: revised_sections/related_work.tex
\section{Related Work}
\label{sec:related_work}

\noindent\textbf{Deep Learning for Time Series Imputation}
Time series imputation has been studied through a wide range of deep learning architectures designed to capture complex temporal dependencies. CNN-based methods extract local temporal patterns, with recent models such as \texttt{TimesNet}~\cite{wu2022timesnet} transforming 1D sequences into 2D representations to capture multi-periodic variations, while \texttt{ModernTCN}~\cite{luo2024moderntcn} employs large-kernel convolutions to enhance temporal feature modeling. Following the success of self-attention, Transformer-based architectures have also been widely adopted for modeling long-range dependencies. For example, \texttt{PatchTST}~\cite{nie2022time} uses patch-level representations to preserve local semantic information, whereas \texttt{SAITS}~\cite{du2023saits} employs diagonal-masked self-attention blocks trained with a joint-optimization strategy to directly target missing entries. In parallel, recent studies show that simpler architectures can also be highly effective. \texttt{DLinear}~\cite{zeng2023transformers} decomposes time series into trend and seasonal components, while \texttt{RLinear}~\cite{li2023revisiting} combines linear mapping with reversible instance normalization to alleviate non-stationary distribution shifts.

Despite these advances, most existing models perform imputation primarily based on the corrupted local context within the input sequence. As a result, they may underutilize informative historical patterns, especially when the observed context is insufficient, weakly correlated, or affected by non-stationary dynamics. In contrast, \framework introduces a retrieval-based framework that explicitly extracts relevant historical subsequences and uses them as auxiliary evidence for reconstructing missing values. This design supplements local-context modeling with long-term historical information, thereby improving reconstruction when the current sequence alone provides limited guidance.

\vspace{3pt}
\noindent\textbf{Retrieval-Augmented Models}
Retrieval-augmented generation (RAG), originally developed in Natural Language Processing, has become an effective paradigm for alleviating the information bottleneck of fixed-parameter models by retrieving relevant instances from external databases~\cite{lewis2020retrieval, gao2023retrieval}. In the temporal domain, recent studies have mainly explored retrieval-augmented forecasting, where similar historical patterns are retrieved to provide long-term context for future trend estimation~\cite{liu2024retrieval,nguyen2026vardiff,han2025retrieval, nguyen2026spectral}. Rather than requiring the model to memorize every possible temporal variation in its parameters, retrieval allows relevant historical sequences to be accessed explicitly, improving the model's ability to handle complex or infrequent patterns.

However, directly applying retrieval to time series imputation is challenging because the query sequence is itself corrupted by missing values. A naive retrieval strategy that matches a corrupted query against complete historical candidates can create a substantial representation mismatch, while dynamically masking and re-encoding every candidate at inference time is computationally expensive. To address this challenge, we introduce \framework, a retrieval-augmented framework tailored for time series imputation. To the best of our knowledge, \framework is among the first frameworks to explicitly formulate deep time series imputation as a retrieval-augmented task, with a dedicated latent alignment mechanism for handling missing-query and complete-candidate mismatch. Its core component, Latent Embedding Alignment, aligns corrupted queries and clean historical candidates within a shared latent space through post-hoc masking and candidate-guided query encoding strategies, enabling both robust retrieval and efficient cached inference.

%% file: revised_sections/methodology.tex
\section{Methodology}
\label{sec:method}

\subsection{Problem Formulation}

Given a complete time series $\mathbf{Y} \in \mathbb{R}^{C \times L}$ with $C$ channels and length $L$, missing values are indicated by a binary mask $\mathbf{M} \in \{0,1\}^{C \times L}$, where $M_{c,l}=1$ means that the entry at channel $c$ and timestamp $l$ is observed, and $M_{c,l}=0$ means that it is missing. The partially observed input sequence is denoted as $\mathbf{X}$, which contains the observed entries of $\mathbf{Y}$ under the mask $\mathbf{M}$.

The goal of time series imputation is to estimate the missing entries of $\mathbf{Y}$ from the incomplete observation $\mathbf{X}$ and the mask $\mathbf{M}$. Specifically, the model is expected to reconstruct the unobserved values indicated by $(\mathbf{1}-\mathbf{M})$ while preserving consistency with the observed values indicated by $\mathbf{M}$.

\subsection{Framework Overview}

\framework extends standard time series imputation by incorporating retrieval-augmented historical references. Given a historical time series repository $\mathbf{S} \in \mathbb{R}^{C \times T}$ from the training split, we construct a candidate pool $\mathcal{D}_{\mathrm{train}}$ by extracting subsequences of length $L$. For each incomplete query $\mathbf{X}$ with mask $\mathbf{M}$, \framework retrieves the top-$k$ most relevant historical subsequences $\mathcal{P}=\mathbf{p}_{i}, i= 1\ldots k$ from $\mathcal{D}_{\mathrm{train}}$ and uses them as auxiliary evidence for reconstruction. The reconstruction is produced by a model-agnostic framework:
\begin{equation}
\bar{\mathbf{X}} = \mathcal{F}_{\theta}(\mathbf{X}, \mathbf{M}, \mathcal{P}).
\end{equation}

The final imputed output preserves the observed values and replaces only the missing entries:
\begin{equation}
\bar{\mathbf{Y}} = \mathbf{M} \odot \mathbf{X} + (\mathbf{1}-\mathbf{M}) \odot \bar{\mathbf{X}},
\end{equation}
where $\odot$ denotes element-wise multiplication.

\begin{figure}[!t]
\centering
\includegraphics[width=1.0\linewidth]{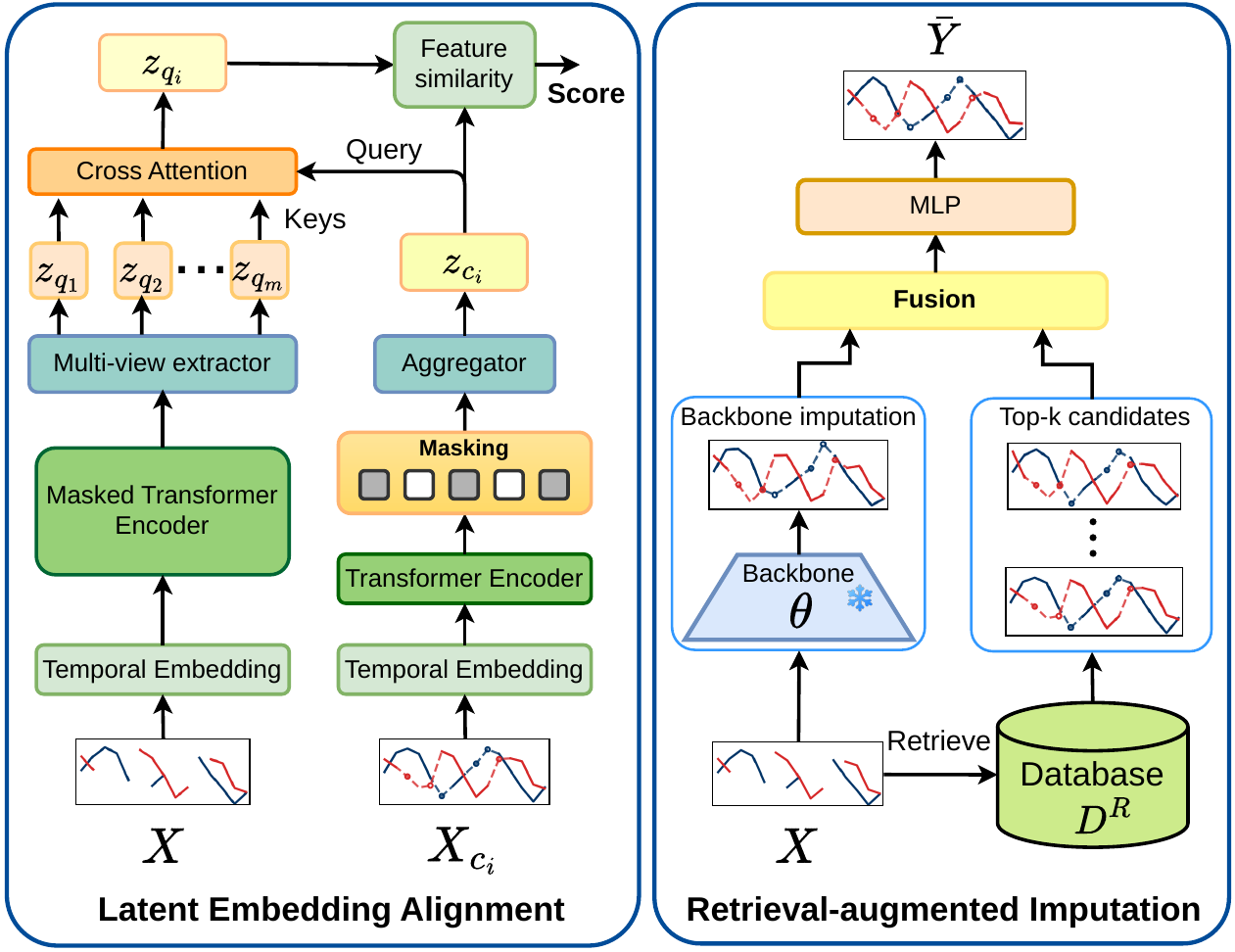}
\caption{\textbf{Overview of the \framework pipeline.} \textbf{(Left)} LEA aligns corrupted queries with clean candidates through late-stage masking and candidate-guided query encoding. \textbf{(Right)} The retrieved historical signals are integrated with the output of a frozen backbone model through lightweight fusion, followed by a MLP-based block for refined reconstruction.}
\label{fig:overview}
\end{figure}

As shown in Fig.~\ref{fig:overview}, \framework follows a two-stage design. In the first stage, we train the Latent Embedding Alignment (LEA) module to align corrupted queries with clean historical candidates in a shared latent space. Historical candidates are encoded independently of query-specific masks and cached as latent representations, enabling efficient retrieval during inference. In the second stage, the retrieved subsequences are integrated with the output of a frozen backbone imputation model through a lightweight adapter. This design allows \framework to supplement local-context reconstruction with relevant historical patterns while remaining compatible with different imputation backbones.

\subsection{Latent Embedding Alignment (LEA)}

\subsubsection{LEA Architecture}
The Latent Embedding Alignment (LEA) module is designed to retrieve structurally relevant historical subsequences for corrupted queries while avoiding the cost of re-encoding the entire database for every query. As illustrated in Fig.~\ref{fig:overview}, LEA adopts a dual-stream architecture consisting of a candidate stream and a query stream. The candidate stream supports offline caching, while the query stream performs mask-aware interaction with cached candidate representations.

\subsubsection{Candidate Encoding}
For each historical candidate $\mathbf{p}_{i} \in \mathcal{D}_{\mathrm{train}}$, LEA first encodes the complete sequence using a Transformer encoder $f_{\theta}$. Importantly, the query-specific mask $\mathbf{M}$ is not applied before candidate encoding. Instead, the candidate is encoded in a mask-agnostic manner, and the interaction with the query mask is postponed to the latent stage. This design allows all candidate representations to be pre-computed and cached offline, avoiding repeated online encoding of the historical database. The cached candidate representation is denoted as
\begin{equation}
\mathbf{z}_{c_{i}}=f_{\theta}(\mathbf{p}_{i}).
\end{equation}

\subsubsection{Candidate-Guided Query Encoding}
For a corrupted query $\mathbf{X}$, LEA extracts multiple contextual views from the observed tokens. Let $\mathcal{U}$ denote the index set of observed query tokens after applying the mask $\mathbf{M}$, and let $\mathbf{h}_{j}, j\in\mathcal{U}$ be their latent representations from the query stream. We introduce $m$ learnable query codes $\mathbf{q}_{r},{r=1 \ldots m}$, where $m \ll L$, to summarize the observed query context:
\begin{equation}
\mathbf{z}_{q_{r}}
=
\sum_{j\in\mathcal{U}} \alpha_{r_{j}}\mathbf{h}_{j},
\quad
\alpha_{r_{j}}
=
\frac{
\exp(\mathbf{q}_{r}^{\top}\mathbf{h}_{j}/\sqrt{d})
}
{
\sum_{u\in\mathcal{U}}
\exp(\mathbf{q}_{r}^{\top}\mathbf{h}_{u}/\sqrt{d})
},
\end{equation}
where $d$ denotes the latent dimension. These $m$ contextual vectors provide multiple views of the observed query. To compare the query with the $i$-th candidate, LEA further aggregates these contextual views using the candidate representation $\mathbf{z}_{c_{i}}$ as guidance:
\begin{equation}
\mathbf{z}_{q_{i}}
=
\sum_{r=1}^{m} \beta_{i_{r}}\mathbf{z}_{q_{r}},
\quad
\beta_{i_{r}}
=
\frac{
\exp(\mathbf{z}_{c_{i}}^{\top}\mathbf{z}_{q_{r}}/\sqrt{d})
}
{
\sum_{v=1}^{m}
\exp(\mathbf{z}_{c_{i}}^{\top}\mathbf{z}_{q_{v}}/\sqrt{d})
}.
\end{equation}

The similarity between the corrupted query and the $i$-th candidate is then computed as
\begin{equation}
s_{i} = \mathbf{z}_{q_i}^{\top}\mathbf{z}_{c_i}.
\end{equation}

Since this interaction is performed only at the top latent layer and uses a small number of query codes, LEA improves representation alignment while maintaining efficient retrieval.

\subsection{Data Construction and Contrastive Training}

\subsubsection{Candidate Pool Generation}
We construct the candidate pool from the training split of the historical repository 
$\mathbf{S}$ by extracting subsequences of length $L$ using a sliding window with 
unit stride, {where $\mathcal{D}_{\mathrm{train}} = \{\mathbf{p}_i\}_{i=1}^{N}$, 
$\mathbf{p}_i = \mathbf{S}_{i:i+L}$, and} $N = T - L + 1$ for a single training series 
of length $T${; for multiple training series, $N$ denotes the total 
number of extracted subsequences.} To reduce distribution shifts across different time periods, each candidate is processed using reversible instance normalization (\texttt{RevIN})~\cite{kim2021reversible} before being encoded and cached.

\subsubsection{Positive Sample Construction}
During training, the complete ground truth $\mathbf{Y}$ is available and is used only to construct supervision for the retriever. To encourage LEA to focus on stable temporal structures rather than stochastic fluctuations, we apply seasonal-trend decomposition (STL)~\cite{cleveland1990stl} to $\mathbf{Y}$:
\begin{equation}
\mathbf{Y}=\mathbf{Y}_{\mathrm{tr}}+\mathbf{Y}_{\mathrm{sea}}+\mathbf{Y}_{\mathrm{res}},
\end{equation}
where $\mathbf{Y}_{\mathrm{tr}}$, $\mathbf{Y}_{\mathrm{sea}}$, and $\mathbf{Y}_{\mathrm{res}}$ denote the trend, seasonal, and residual components, respectively. The positive reference is constructed from the deterministic components:
\begin{equation}
\mathbf{p}^{+}=\mathbf{Y}_{\mathrm{tr}}+\mathbf{Y}_{\mathrm{sea}}.
\end{equation}

This encourages the retriever to align corrupted observations with the underlying structural pattern of the target sequence.

\subsubsection{Hard Negative Mining}
Selecting informative negatives is important because randomly sampled subsequences may be too dissimilar and provide weak training signals. We therefore evaluate the Pearson correlation between the ground truth $\mathbf{Y}$ and each candidate $\mathbf{p}_{i} \in \mathcal{D}_{\mathrm{train}}$:
\begin{equation}
\rho_{i} = \mathrm{Pearson}(\mathbf{Y},\mathbf{p}_{i}).
\end{equation}

High-correlation but non-positive candidates are sampled as hard negatives. These samples are close enough to challenge the retriever but do not correspond to the target structural reference, encouraging LEA to learn fine-grained discriminative features beyond simple linear similarity.

\subsubsection{Contrastive Objective}
Given a corrupted query, its positive reference $\mathbf{p}^{+}$, and a set of hard negatives $\mathcal{N}^{-}$, LEA is trained with an InfoNCE-style contrastive objective:
\begin{equation}
\mathcal{L}_{\mathrm{LEA}}
=
-\log
\frac{
\exp(s(\mathbf{X},\mathbf{p}^{+})/\tau)
}
{
\exp(s(\mathbf{X},\mathbf{p}^{+})/\tau)
+
\displaystyle\sum_{\mathbf{p}^{-}\in\mathcal{N}^{-}}
\exp(s(\mathbf{X},\mathbf{p}^{-})/\tau)
},
\end{equation}
where $s(\cdot,\cdot)$ denotes the LEA similarity function and $\tau$ is a temperature parameter. The same missingness pattern $\mathbf{M}$ is applied through the post-hoc masking procedure when comparing the query with positive and negative samples, ensuring that similarities are computed in an aligned latent space.

\subsection{Retrieval-Augmented Time Series Imputation}

After LEA is trained, \framework uses it to retrieve relevant historical subsequences and integrate them with the output of a backbone imputation model.

\subsubsection{Retrieval Module}
We first build a persistent latent index by pre-computing candidate representations:
\begin{equation}
\mathcal{D}^{V}
=
\mathbf{z}_{c_{i}}=f_{\theta}(\mathbf{p}_{i})
\end{equation}
For a corrupted query $\mathbf{X}$ with mask $\mathbf{M}$, LEA computes similarity scores $s_{i}$ with $i = 1 \ldots N$ between the query and all cached candidates. The top-$k$ candidates are selected as retrieved references:
\begin{equation}
\mathcal{K}
=
\operatorname{Top-k}({s_i}),
\quad
\mathcal{P}
=
{\mathbf{p}_{i} \mid i \in \mathcal{K}},
\end{equation}
where $\operatorname{Top-k}(\cdot)$ returns the indices of the $k$ largest similarity scores. Because candidate embeddings are cached offline, retrieval only requires query encoding and lightweight latent-space matching during inference.

\subsubsection{Lightweight Adapter}
The retrieved candidates $\mathcal{P}$ are integrated with the prediction from a frozen backbone imputation model. Let $\mathbf{z}_{\mathrm{bb}}$ denote the backbone output representation. To mitigate distribution shifts between the current query and retrieved references, we apply RevIN to both $\mathbf{z}_{\mathrm{bb}}$ and the retrieved candidates. The normalized candidates are then aggregated into a retrieval representation $\mathbf{z}_{\mathrm{re}}$ by average pooling over the top-$k$ references.

To adaptively combine the backbone output with the retrieval representation, we compute a gating matrix:
\begin{equation}
\mathbf{g}
=
\sigma(\mathrm{MLP}_g(\mathbf{z}_{\mathrm{bb}})),
\end{equation}
where $\sigma(\cdot)$ denotes the sigmoid activation. The fused representation is obtained as
\begin{equation}
\mathbf{z}_{f}
=
\mathbf{g} \odot \mathbf{z}_{\mathrm{bb}}
+
(\mathbf{1}-\mathbf{g})\odot \mathbf{z}_{\mathrm{re}}.
\end{equation}
The fused representation is further refined by a residual MLP:
\begin{equation}
\mathbf{z}_{\mathrm{imp}}
=
\mathbf{z}_{f}
+
\mathrm{MLP}_{r}(\mathbf{z}_{f}).
\end{equation}
Finally, the denormalization step restores $\mathbf{z}_{\mathrm{imp}}$ to the original scale and produces the reconstruction $\bar{\mathbf{X}}$. Since the backbone is frozen, the additional trainable parameters introduced by \framework mainly come from the two compact MLPs in the adapter, making the framework lightweight and easy to integrate with different imputation backbones.

\subsubsection{Imputation Training Objective}
The adapter is optimized using the mean squared error over the missing entries:
\begin{equation}
\mathcal{L}_{\mathrm{imp}}
=
\frac{
|(\mathbf{1}-\mathbf{M})\odot(\mathbf{Y}-\bar{\mathbf{Y}})|_{F}^{2}
}
{
|\mathbf{1}-\mathbf{M}|_{1}
},
\end{equation}
where $|\cdot|_{F}$ and $|\cdot|_{1}$ denote the Frobenius norm and $\ell_{1}$ norm, respectively. This objective ensures that learning focuses directly on the missing entries while the final output keeps the observed values unchanged.

%% file: revised_sections/experiments.tex
\section{Experiments}

\subsection{Experimental Setup}

\subsubsection{Datasets}
Following established time series benchmarks~\cite{zhou2021informer,wu2022timesnet,luo2024moderntcn}, we evaluate \framework on six widely used real-world datasets. The \textbf{ETT} benchmark~\cite{zhou2021informer} contains four subsets, \textbf{ETTh1}, \textbf{ETTh2}, \textbf{ETTm1}, and \textbf{ETTm2}, which record electricity-transformer oil temperature and load at hourly or 15-minute granularity. \textbf{Electricity}~\cite{dua2017uci} contains hourly electricity consumption from 321 clients over three years. \textbf{Weather}~\cite{zhou2021informer} contains 21 meteorological variables recorded every 10 minutes throughout 2020. Together, these datasets cover diverse temporal patterns, sampling granularities, and distributional changes. For all datasets, we follow the same preprocessing and train/validation/test splits used by the baseline benchmarks to ensure a fair comparison.

\subsubsection{Baseline Methods}
To evaluate whether \framework can serve as a model-agnostic enhancement, we instantiate it with seven general-purpose backbones from three architectural families: CNN-based models, including \texttt{ModernTCN}~\cite{luo2024moderntcn} and \texttt{TimesNet}~\cite{wu2022timesnet}; Transformer-based models, including \texttt{PatchTST}~\cite{nie2022time}, \texttt{Crossformer}~\cite{zhang2023crossformer}, and \texttt{Autoformer}~\cite{wu2021autoformer}; and Linear/MLP-based models, including \texttt{DLinear}~\cite{zeng2023transformers} and \texttt{RLinear}~\cite{li2023revisiting}. These backbones cover a broad range of temporal modeling mechanisms, including multi-scale convolution, patch-wise attention, decomposition, and linear temporal projection. We compare each backbone with its \framework-augmented counterpart to isolate the benefit of retrieval-enhanced historical context. We further include three imputation-specialized baselines: \texttt{SAITS}~\cite{du2023saits} and \texttt{Helix}~\cite{zhang2026helix} are dedicated imputation architectures, while \texttt{Glocal-IB}~\cite{yang2026glocal} is a concurrent model-agnostic training paradigm that augments backbone objectives with a global alignment loss derived from the Information Bottleneck framework. This separation allows us to assess \framework across both general-purpose and imputation-specialized settings.

\subsubsection{Implementation Details}
We train LEA with a batch size of 16. The query set contains $m=16$ codes, each represented by a 64-dimensional vector. The learning rate is set to $0.001$. The temperature parameter is initialized as $\tau=\ln(1/0.07)$ and is learned during training. This initialization yields a sufficiently sharp initial softmax distribution for the contrastive objective while allowing the scaling factor to adapt to each dataset.

We integrate \framework with each backbone and optimize the adapter using Adam and the MSE loss. For fairness, we keep the official training pipeline and hyperparameter configurations of each backbone unchanged, consulting Time-Series-Library\footnote{\url{https://github.com/thuml/Time-Series-Library}} when needed. Thus, any performance difference between a backbone and its augmented version can be attributed to \framework rather than backbone-specific retuning. All experiments are conducted on a single NVIDIA GeForce RTX 5090 GPU. The source code is available at \url{https://anonymous.4open.science/r/Time-series-0142/}.

\subsubsection{Evaluation}
We use MSE and MAE as evaluation metrics. To assess robustness under different levels of information loss, we evaluate four missing rates $r\in\{0.125,0.25,0.375,0.5\}$ and four input lengths $L\in\{96,192,336,720\}$. Each configuration is repeated with three random seeds, and we report the average result.

\subsection{Overall Results}

\input{tables/promotion}

Table~\ref{tab:promotion} summarizes the relative MSE improvement obtained by adding \framework to different backbones across all datasets and missing rates. Overall, \framework improves nearly all evaluated configurations, with only one marginal degradation. This demonstrates that the proposed retrieval-augmented mechanism is broadly effective rather than being tied to a specific architecture or dataset.

Several observations can be made. First, the gains are particularly clear for linear backbones such as \texttt{DLinear} and \texttt{RLinear}. This suggests that explicit historical retrieval can compensate for the limited capacity of simple parametric models by providing useful non-parametric context. Second, \texttt{Transformer}-based models also benefit substantially, indicating that retrieval remains helpful even when the backbone already has strong temporal modeling ability. Third, recent competitive architectures such as \texttt{ModernTCN} and \texttt{TimesNet} still obtain consistent improvements, showing that \framework can enhance strong local-context models without modifying their internal structures.

We also observe that \framework is compatible with imputation-specialized methods and concurrent model-agnostic training paradigms. In particular, applying \framework on top of \texttt{Glocal-IB}-trained backbones still brings further improvement, suggesting that retrieval-based context enrichment and objective-level regularization are complementary. These results support the main claim that \framework serves as a general plug-and-play module for time series imputation. Full MSE and MAE results are reported in Table~\ref{tab:full_results_mse} and Table~\ref{tab:full_results_mae}, respectively.

\subsection{Model Analysis}

\subsubsection{Influence of Retrieval Mechanisms}

\input{tables/ablation_retrieval}

We first examine whether the observed gains come from meaningful retrieval rather than simply adding extra historical sequences. Table~\ref{tab:ablation-retrieval} compares \framework with random retrieval, metric-based retrieval, and two embedding-based alternatives. Random retrieval provides little benefit and can even hurt performance, showing that retrieval quality is critical for time-series imputation. Metric-based criteria, such as Pearson correlation and DTW after linear interpolation, provide stronger candidates than random retrieval but remain limited because they mainly capture surface-level similarity.

Among embedding-based methods, AR supports offline candidate encoding but suffers from a representation mismatch between masked queries and complete historical candidates. OM reduces this mismatch by encoding candidates under the query-specific mask, but this requires online candidate re-encoding and therefore becomes expensive when the retrieval database is large. LEA offers a more favorable trade-off: it preserves offline candidate pre-computation while using latent interaction to align masked queries with historical candidates. As a result, \framework achieves performance comparable to or better than OM, while avoiding its online encoding bottleneck.

\subsubsection{Importance of Candidate-Guided Query Encoding}

\input{tables/interaction_ablation}

As illustrated in Fig.~\ref{fig:overview} (Left), we evaluate the candidate-guided query encoding mechanism in LEA through an ablation study against a Bi-Encoder baseline. This baseline retains the late-stage masking pipeline but restricts the query representation, thereby collapsing the dynamic cross-attention process into non-interactive embedding alignment.

As summarized in Table~\ref{tab:interaction_ablation}, removing this mechanism consistently degrades performance across all benchmarks, with especially clear drops on \textbf{Electricity} and \textbf{ETTm1}. These results confirm that jointly encoding the masked query with candidate information helps extract richer contextual representations, leading to more accurate imputation.

We further place the contextual views on the input query branch rather than the candidate branch. If the multi-view representation were deployed on the candidate side, the cached representation of each candidate would need to pass through multiple attention layers and an additional cross-attention block during online candidate encoding. This design would substantially increase inference latency, especially when scaling to large historical databases.

\subsubsection{Effectiveness of \framework's Fusion}

To assess the design of \framework's Fusion (AF) mechanism, we compare it with two alternative integration strategies on \texttt{Crossformer} under $r=0.5$: (\textit{i})~\textit{Linear Fusion}, which projects the concatenation of the backbone output and retrieved references through a static linear layer; and (\textit{ii})~\textit{Cross-attention Fusion}, which computes attention weights to dynamically align the backbone output with the top-$k$ retrieved candidates.

As shown in Table~\ref{tab:fusion_ablation}, AF consistently achieves the lowest MSE across the evaluated datasets. Linear Fusion, despite its simplicity, relies on static parameterization and cannot sufficiently capture sample-specific retrieval patterns. Cross-attention Fusion is more expressive, but it introduces additional parameter overhead and may overfit, notably degrading performance on Electricity compared with the standalone backbone. In contrast, AF avoids both limitations through a lightweight, instance-dependent gating mechanism that dynamically calibrates the retrieved context against the backbone output with minimal additional parameters.

\input{tables/fusion_ablation}

\subsubsection{Ablation Study on Component Contributions}

\input{tables/ablation_components}

We investigate the individual contributions of two key components in \framework: the gating matrix $g$ and the \texttt{RevIN} module for mitigating distribution shift. As summarized in Table~\ref{tab:ablation-components}, removing either component consistently degrades performance and, in some cases, even performs worse than the vanilla backbone. Specifically, the gating mechanism acts as a dynamic filter that selectively integrates retrieved historical patterns, preventing irrelevant context from distorting the current reconstruction. \texttt{RevIN} further mitigates distribution shift between historical references and the current input, enabling stable and scale-invariant imputation under non-stationary temporal dynamics.

\subsubsection{Influence of the Number of Retrieved Candidates $k$}
\label{subsec:k_sensitivity_analysis}

We analyze the sensitivity of imputation performance to the number of retrieved candidates $k \in {1, 2, 3, 5, 10}$ across six benchmark datasets, as shown in Table~\ref{tab:appendix_k_sensitivity}. \framework remains stable across the evaluated range, reducing the need for exhaustive hyperparameter search. While a small value such as $k=1$ is sufficient for datasets with simpler temporal patterns, it can be insufficient for datasets with more complex dynamics, where richer contextual information is beneficial. Conversely, excessively large values may introduce noise from less relevant candidates and slightly degrade performance. Overall, the best results are typically achieved around $k \in {2,3}$. We therefore fix $k=3$ as the default setting for all experiments.


\input{tables/k_analysis}

\subsubsection{Theoretical Complexity Analysis}

\input{tables/complexity_comparison_no_c}

To characterize the computational cost of LEA, we compare its theoretical complexity with the embedding-based retrieval paradigms introduced in Fig.~\ref{fig:intro_image}, namely AR and OM, across different operational phases.

Table~\ref{tab:complexity_comparison_no_c} summarizes the theoretical bounds. Given a missing rate $r$, where $r'=1-r$ denotes the observed ratio, a standard Transformer backbone processing the observed tokens requires $\mathcal{O}(L^2 r'^2 d + L r' d^2)$ operations per block, accounting for both quadratic self-attention and latent linear projections.

The OM strategy incurs high online inference cost because its dynamic coupling mechanism requires the query mask to guide the encoding of historical candidates. As a result, candidate representations must be recomputed online, leading to an overhead of $\mathcal{O}(N(L^2 r'^2 d + L r' d^2))$, which scales linearly with the database size $N$ and quadratically with the observed sequence length $Lr'$.

LEA mitigates this bottleneck through post-hoc masking. The expensive sequence encoding can be shifted to the offline phase, with complexity $\mathcal{O}(N(L^2 d + Ld^2))$. During online retrieval, LEA reduces candidate encoding to $\mathcal{O}(NLr'd)$. Although LEA introduces an additional similarity-estimation cost of $\mathcal{O}(Nm(Lr'+1)d)$ due to candidate-guided query encoding, this term remains efficient because the number of query codes is small, i.e., $m \ll L$. Therefore, LEA preserves the efficiency of offline-indexed retrieval while improving representation consistency under dynamic masking.

\subsubsection{Inference Latency and Empirical Scalability}

\input{tables/efficiency_analysis}

While \framework reduces training cost by freezing the backbone parameters and optimizing only a lightweight adapter, its retrieval mechanism introduces a small inference overhead. As reported in Table~\ref{tab:efficiency-analysis}, this overhead is consistently limited across the evaluated architectures and remains small in absolute terms. This observation is consistent with the theoretical analysis in Table~\ref{tab:complexity_comparison_no_c}. The additional cost mainly comes from candidate-guided query encoding and similarity matching against the pre-computed candidate embedding database.

Importantly, the latency remains bounded in practice. The post-hoc masking and aggregation design of LEA restricts online candidate encoding to $\mathcal{O}(N L r' d)$, avoiding the quadratic cost of conventional dynamic retrieval approaches. Even for lightweight backbones, where the relative overhead is more visible, the absolute inference time remains modest. These results show that \framework achieves a favorable trade-off between retrieval-augmented accuracy and computational efficiency.

\subsubsection{\framework Remains Helpful under Temporal Distribution Shift}

\input{tables/training_data_proportion}

We evaluate the resilience of \framework under temporal distribution shift by restricting the training data to 25\% and 50\% of the full training set. Since these datasets are chronologically ordered, limiting the training window to earlier segments creates a challenging setting where the retrieval database is temporally separated from the test period.

As shown in Table~\ref{tab:training_data_proportion}, \framework matches or outperforms the standalone \texttt{PatchTST} baseline across all datasets and data regimes, with the only tie occurring on \textbf{Weather} under the 25\% setting. Even in the most constrained setting, LEA recovers useful structural patterns from limited historical memory. Under the 50\% setting, \framework continues to provide consistent improvements. With the full training set, the gains become more pronounced, suggesting that the framework can effectively leverage a denser retrieval repository to improve imputation accuracy.

\input{tables/infrequent_patterns}

\subsubsection{Robustness to Infrequent Temporal Patterns}

\input{tables/full_results_mse}
\input{tables/full_results_mae}

Parametric deep networks can underrepresent long-tail temporal behaviors because their parameters are primarily optimized toward dominant global trajectories. \framework addresses this limitation through an explicit non-parametric retrieval mechanism, enabling the reconstruction of localized dynamics even when their morphological signatures are rare. We validate this property on synthetic time series embedded with sparse, event-driven anomalies.

\paragraph{Synthetic Data Generation}
The evaluation sequences superimpose deterministic global trajectories, including sinusoidal trend and seasonality, with localized low-frequency structural anomalies modeled by a Time-Varying Autoregressive (\texttt{TV-AR})~\cite{bringmann2017changing} process:
\[
x_t = \sum_{i=1}^{p} \varphi_i(t), x_{t-i} + \epsilon_t,
\]
where $p=30$, $\epsilon_t \sim \mathcal{U}(-0.05, 0.05)$, and $\varphi_i(t)$ changes abruptly to define the rare anomaly morphology within a fixed window of 192 time steps. To enforce data scarcity, \texttt{TV-AR} segments are injected into the training set with exactly $1$, $2$, and $4$ occurrences.

\paragraph{Empirical Evaluation}

Table~\ref{tab:infrequent_patterns} reports the imputation performance. Standalone parametric baselines degrade as the pattern frequency decreases from 4 to 1, indicating their difficulty in modeling rare temporal patterns. In contrast, \framework mitigates this degradation across all backbones, and its relative gains become larger as the pattern becomes scarcer. For example, \framework reduces \texttt{DLinear}'s MSE by $14.1\%$ with one occurrence, compared with $5.3\%$ with four occurrences. These results confirm that non-parametric retrieval provides explicit temporal context that helps overcome the data-scarcity bottleneck of purely parametric models.

\subsection{Detailed Results}

Tables~\ref{tab:full_results_mse} and~\ref{tab:full_results_mae} report the full MSE and MAE results for all ten baselines and their \framework-augmented counterparts across six datasets, four missing rates, and four sequence lengths, confirming the consistent gains summarized in Table~\ref{tab:promotion}.

%% file: tables/promotion.tex
\begin{table*}[t]
\centering
\footnotesize
\renewcommand{\arraystretch}{1.3}
\setlength{\tabcolsep}{5pt}
\caption{MSE improvement (\%) obtained by adding \framework to ten baselines on six datasets. Higher values indicate larger improvements. Results are averaged over four input lengths $L\in\{96,192,336,720\}$ for each missing rate $r\in\{0.125,0.25,0.375,0.5\}$.}
\label{tab:promotion}
\begin{tabular}{p{0.9cm}p{0.5cm} *{3}{>{\centering\arraybackslash}p{1.25cm}}|*{7}{>{\centering\arraybackslash}p{1.3cm}}}
\cline{1-12}
\multicolumn{2}{c}{Baseline type} &
\multicolumn{3}{c|}{\textbf{Imputation-specialized}} &
\multicolumn{7}{c}{\textbf{General-purpose}} \\ \cline{3-12}
\textbf{Dataset} & $r$ &
\texttt{Helix} & \texttt{SAITS} & \texttt{Glocal-IB} & \texttt{ModernTCN} & \texttt{TimesNet} & \texttt{Crossformer} & \texttt{Autoformer} & \texttt{PatchTST} & \texttt{DLinear} & \texttt{RLinear} \\
\hline
\multirow{4}{*}{\textbf{ETTh1}}
 & 0.125 & 10.26 & 19.91 & 13.05 & 8.11  & 6.90  & 49.14 & 24.62 & 35.90 & 33.02 & 27.10 \\
 & 0.25  & 0.00         & 7.93  & 8.63  & 10.87 & 8.90  & 41.67 & 27.71 & 33.85 & 33.57 & 27.34 \\
 & 0.375 & 4.73  & 7.83  & 10.90 & 5.17  & 10.60 & 38.00 & 24.53 & 27.46 & 31.03 & 24.28 \\
 & 0.5   & 5.59  & 12.36 & 11.01 & 8.22  & 12.80 & 28.82 & 21.71 & 25.45 & 28.10 & 21.63 \\ \hline
\multirow{4}{*}{\textbf{ETTh2}}
 & 0.125 & 5.50  & 9.56  & 15.98 & 2.50  & 5.10  & 29.51 & 8.33  & 3.13  & 24.53 & 14.91 \\
 & 0.25  & 2.44  & 13.13 & 10.54 & 2.27  & 6.30  & 26.62 & 9.26  & 1.45  & 26.09 & 24.06 \\
 & 0.375 & 3.19  & 8.45  & 11.05 & 2.08  & 7.10  & 21.52 & 6.56  & 4.05  & 27.22 & 25.16 \\
 & 0.5   & 1.63  & 12.07 & 6.84  & 8.77  & 8.30  & 17.49 & 4.05  & 1.25  & 24.50 & 23.78 \\ \hline
\multirow{4}{*}{\textbf{ETTm1}}
 & 0.125 & 6.98  & 23.46 & 9.05  & 10.53 & 2.50  & 21.15 & 44.44 & 41.79 & 43.40 & 43.40 \\
 & 0.25  & 4.00  & 15.52 & 6.90  & 12.50 & 3.40  & 20.37 & 40.98 & 42.42 & 39.44 & 40.00 \\
 & 0.375 & 7.02  & 23.81 & 8.16  & 15.38 & 3.60  & 13.56 & 49.37 & 43.84 & 34.07 & 34.83 \\
 & 0.5   & 12.07 & 18.95 & 5.22  & 7.14  & 4.10  & 15.63 & 46.59 & 34.94 & 20.19 & 44.77 \\ \hline
\multirow{4}{*}{\textbf{ETTm2}}
 & 0.125 & 14.29 & 20.73 & 18.06 & 0.00 & 2.00  & 16.42 & 24.59 & 6.25  & 21.88 & 19.67 \\
 & 0.25  & 0.00 & 15.38 & 15.80 & 4.76  & 2.20  & 17.50 & 24.62 & 5.71  & 22.62 & 20.25 \\
 & 0.375 & 10.53 & 15.44 & 6.92  & 4.00  & 2.40  & 13.95 & 8.89  & 2.70  & 13.33 & 17.89 \\
 & 0.5   & 8.70  & 10.62 & 6.78  & -3.85 & 2.70  & 12.00 & 6.67  & 7.50  & 20.47 & 20.54 \\ \hline
\multirow{4}{*}{\textbf{Electricity}}
 & 0.125 & 12.75 & 7.35  & 7.89  & 7.14  & 9.80  & 10.29 & 6.52  & 10.99 & 17.46 & 19.05 \\
 & 0.25  & 3.70  & 8.33  & 5.89  & 8.33  & 9.50  & 11.88 & 7.97  & 5.68  & 19.42 & 18.09 \\
 & 0.375 & 3.31  & 11.06 & 5.70  & 3.70  & 9.70  & 11.30 & 8.28  & 5.32  & 17.19 & 17.39 \\
 & 0.5   & 3.17  & 12.27 & 2.71  & 6.74  & 10.00 & 16.28 & 8.61  & 1.83  & 18.06 & 15.94 \\ \hline
\multirow{4}{*}{\textbf{Weather}}
 & 0.125 & 14.81 & 6.90  & 13.29 & 10.34 & 2.70  & 10.53 & 18.75 & 14.71 & 15.79 & 15.79 \\
 & 0.25  & 6.45  & 10.34 & 6.08  & 3.33  & 3.20  & 5.26  & 16.22 & 23.91 & 19.15 & 19.15 \\
 & 0.375 & 11.76 & 11.11 & 6.96  & 3.03  & 3.60  & 7.14  & 12.12 & 15.38 & 23.64 & 22.22 \\
 & 0.5   & 10.53 & 8.33  & 7.24  & 0.00  & 3.80  & 9.09  & 16.00 & 27.78 & 21.88 & 22.58 \\
\bottomrule
\end{tabular}
\end{table*}

%% file: tables/ablation_retrieval.tex
\begin{table}[ht]
\centering
\renewcommand{\arraystretch}{1.6}
\setlength{\tabcolsep}{2.5pt}
\caption{Ablation study of retrieval criteria on \textbf{ETTh1} and \textbf{Weather} with missing rate $r=0.25$. Results are averaged over four input lengths $L\in\{96,192,336,720\}$. The entry ``-'' denotes the vanilla baseline without retrieval augmentation, while other entries represent the specific methods used for candidate selection. Best
results are in \textbf{bold}.}
\label{tab:ablation-retrieval}
\resizebox{\columnwidth}{!}{
\begin{tabular}{cccccccccc}
\toprule
\multicolumn{2}{c}{\textbf{Dataset}} &
  \multicolumn{4}{c}{\textbf{ETTh1}} &
  \multicolumn{4}{c}{\textbf{Weather}} \\ \midrule
\multicolumn{2}{c}{Model} &
  \multicolumn{2}{c}{\texttt{ModernTCN}} &
  \multicolumn{2}{c}{\texttt{DLinear}} &
  \multicolumn{2}{c}{\texttt{ModernTCN}} &
  \multicolumn{2}{c}{\texttt{DLinear}} \\ 
\multicolumn{2}{c}{} &
  \textbf{MAE} &
  \multicolumn{1}{c}{\textbf{MSE}} &
  \textbf{MAE} &
  \textbf{MSE} &
  \textbf{MAE} &
  \multicolumn{1}{c}{\textbf{MSE}} &
  \textbf{MAE} &
  \textbf{MSE} \\ \hline
\multicolumn{1}{c}{\multirow{2}{*}{}} &
  - &
  0.149 &
  \multicolumn{1}{c}{0.046} &
  0.261 &
  0.140 &
  0.058 &
  \multicolumn{1}{c}{0.030} &
  0.107 &
  0.047 \\
\multicolumn{1}{c}{} &
  Random &
  0.158 &
  \multicolumn{1}{c}{0.052} &
  0.258 &
  0.136 &
  0.067 &
  \multicolumn{1}{c}{0.035} &
  0.113 &
  0.048 \\ \hline
\multicolumn{1}{c}{\multirow{2}{*}{\begin{tabular}[c]{@{}c@{}}\textbf{Metric} \\ \textbf{based}\end{tabular}}} &
  Pearson &
  0.135 &
  \multicolumn{1}{c}{0.042} &
  0.218 &
  0.099 &
  0.059 &
  \multicolumn{1}{c}{0.030} &
  0.096 &
  0.043 \\
\multicolumn{1}{c}{} &
  DTW &
  0.146 &
  \multicolumn{1}{c}{0.044} &
  0.244 &
  0.127 &
  0.058 &
  \multicolumn{1}{c}{0.032} &
  0.093 &
  0.045 \\ \hline
\multicolumn{1}{c}{\multirow{3}{*}{\begin{tabular}[c]{@{}c@{}}\textbf{Embedding}\\ \textbf{based}\end{tabular}}} &
  AR &
  0.136 &
  \multicolumn{1}{c}{0.044} &
  0.217 &
  0.103 &
  0.060 &
  \multicolumn{1}{c}{0.032} &
  0.094 &
  0.042 \\
\multicolumn{1}{c}{} &
  OM &
  \textbf{0.132} &
  \multicolumn{1}{c}{\textbf{0.041}} &
  {\underline{0.214}} &
  {\underline{0.094}} &
  \textbf{0.054} &
  \multicolumn{1}{c}{\textbf{0.029}} &
  {\underline{0.085}} &
  {\underline{0.040}} \\
\multicolumn{1}{c}{} &
  \framework &
  {\underline{0.133}} &
  \multicolumn{1}{c}{\textbf{0.041}} &
  \textbf{0.211} &
  \textbf{0.093} &
  {\underline{0.056}} &
  \multicolumn{1}{c}{\textbf{0.029}} &
  \textbf{0.084} &
  \textbf{0.038} \\ \bottomrule
\end{tabular}
}
\end{table}

%% file: tables/interaction_ablation.tex
\begin{table}[t!]
\centering
\caption{Ablation study of the candidate-guided query encoding in MSE.}
\label{tab:interaction_ablation}
\footnotesize
\setlength{\tabcolsep}{6pt}
\renewcommand{\arraystretch}{1.2}
\begin{tabular}{@{}cccc@{}}
\toprule
\textbf{Dataset} & \textbf{w/o Interaction} & \framework & \textbf{Prm. ($\Delta\%$)} \\ 
\midrule
\textbf{ETTh1}       & 0.128 & \textbf{0.123} & 3.91\%  \\
\textbf{ETTh2}       & 0.082 & \textbf{0.079} & 3.66\%  \\
\textbf{ETTm1}       & 0.068 & \textbf{0.060} & 11.76\% \\
\textbf{ETTm2}       & 0.040 & \textbf{0.037} & 7.50\%  \\
\textbf{Weather}     & 0.109 & \textbf{0.107} & 1.83\%  \\
\textbf{Electricity} & 0.046 & \textbf{0.039} & 15.22\% \\ 
\bottomrule
\end{tabular}
\end{table}

%% file: tables/fusion_ablation.tex
\begin{table}[t!]
\centering
\caption{Ablation study evaluating different fusion strategies (MSE). Bold indicates the optimal configuration. ``+\texttt{Cross-attn}'' denotes the cross-attention fusion and ``+AF'' denotes our \framework's Fusion. }
\label{tab:fusion_ablation}
\footnotesize
\setlength{\tabcolsep}{6pt}
\renewcommand{\arraystretch}{1.2}
\begin{tabular}{@{}ccccc@{}}
\toprule
\textbf{Dataset} & {\texttt{Crossformer}} & \texttt{+Linear} & \texttt{+Cross-attn} & {+AF} \\ 
\midrule
\textbf{ETTh1}       & 0.170 & 0.156 & 0.158 & \textbf{0.121} \\
\textbf{ETTm1}       & 0.064 & 0.063 & 0.061 & \textbf{0.054} \\
\textbf{Weather}     & 0.129 & 0.117 & 0.112 & \textbf{0.108} \\
\textbf{Electricity} & 0.044 & 0.046 & 0.048 & \textbf{0.043} \\ 
\bottomrule
\end{tabular}
\end{table}

%% file: tables/ablation_components.tex
\begin{table}[t!]
\centering
\caption{Ablation study on different components of the \framework mechanism. We evaluate the impact of the gating module, \texttt{RevIN}, and the baseline imputation on final performance.}
\label{tab:ablation-components}
\footnotesize
\setlength{\tabcolsep}{10pt}
\renewcommand{\arraystretch}{1.2}
\begin{tabular}{@{}ccccc@{}}
\toprule
{Dataset} & \multicolumn{2}{c}{{\textbf{ETTh1}}} & \multicolumn{2}{c}{{\textbf{Weather}}} \\
\midrule
{Metric} & {MAE} & {MSE} & {MAE} & {MSE} \\ 
\midrule
\texttt{Autoformer} & 0.211 & 0.097 & 0.126 & 0.051 \\ 

{\texttt{Autoformer} +\framework} & \textbf{0.191} & \textbf{0.081} & \textbf{0.102} & \textbf{0.040} \\
{w/o} gate $g$ & 0.202 & 0.095 & 0.118 & 0.053 \\
{w/o} \texttt{RevIN} & 0.209 & 0.098 & 0.121 & 0.049 \\ 
\bottomrule
\end{tabular}
\end{table} 

%% file: tables/k_analysis.tex
\begin{table}[t]
\centering
\caption{Detailed MSE results of sensitivity analysis under varying numbers of retrieved candidates ($k$).}
\label{tab:appendix_k_sensitivity}
\footnotesize
\setlength{\tabcolsep}{10pt}
\renewcommand{\arraystretch}{1.2}
\begin{tabular}{@{}cccccc@{}}
\toprule
\textbf{Dataset} & $k$=1 & $k$=2 & $k$=3 & $k$=5 & $k$=10 \\ 
\midrule
\textbf{ETTh1}       & 0.124 & 0.123 & \textbf{0.123} & 0.125 & 0.124 \\
\textbf{ETTh2}       & \textbf{0.078} & 0.081 & 0.079 & 0.081 & 0.082 \\
\textbf{ETTm1}       & 0.062 & 0.062 & \textbf{0.060} & 0.063 & 0.061 \\
\textbf{ETTm2}       & 0.040 & 0.040 & \textbf{0.037} & \textbf{0.037} & 0.039 \\
\textbf{Weather}     & 0.107 & 0.109 & \textbf{0.107} & 0.110 & 0.114 \\
\textbf{Electricity} & 0.041 & \textbf{0.038} & 0.039 & 0.041 & 0.043 \\ 
\bottomrule
\end{tabular}
\end{table}

%% file: tables/complexity_comparison_no_c.tex
\begin{table*}[ht]
\centering
\caption{Theoretical complexity comparison of retrieval strategies for time-series imputation. $N$: database size, $L$: sequence length, $d$: latent dimension, $m$: number of query codes ($m \ll L$), and $r'=1-r$, where $r$ denotes the missing rate.}
\label{tab:complexity_comparison_no_c}
\renewcommand{\arraystretch}{1.6}
\footnotesize
\setlength{\tabcolsep}{5pt}
\begin{tabular}{lccc}
\toprule
\textbf{Phase / Strategy} & \textbf{AR} & \textbf{OM} & \textbf{LEA (Ours)} \\
\midrule
{Offline Embedding} & $\mathcal{O}(N(L^2 d + L d^2))$ & 0 (Cannot) & $\mathcal{O}(N(L^2 d + L d^2))$ \\
{Online Query Encoding} & $\mathcal{O}(L^2 r'^2 d + L r' d^2)$ & $\mathcal{O}(L^2 r'^2 d + L r' d^2)$ & $\mathcal{O}(L^2 r'^2 d + L r' d^2)$ \\
{Online Candidate Encoding} & $\mathcal{O}(1)$ & $\mathcal{O}(N(L^2 r'^2 d + L r' d^2))$ & $\mathcal{O}(N L r' d)$ \\
{Similarity Estimation} & $\mathcal{O}(N d)$ & $\mathcal{O}(N d)$ & $\mathcal{O}(N m (L r' + 1)d)$ \\
\bottomrule
\end{tabular}
\end{table*}

%% file: tables/efficiency_analysis.tex
\begin{table}[ht]
\centering
\renewcommand{\arraystretch}{1.2}
\caption{Efficiency comparison of baseline models and their \framework-enhanced versions on Weather under $r=0.375$ and $L=192$. Best results are \textbf{bolded}.}
\label{tab:efficiency-analysis}
\footnotesize
\begin{tabular}{lp{1.2cm}p{1.2cm}p{1.2cm}p{1.2cm}}
\hline
\textbf{Model} & \textbf{Trainable Params} & \textbf{Train Time (s)} & \textbf{Const. RDB Time (s)} & \textbf{Inference Time (ms)} \\ \hline
\texttt{ModernTCN} & 5.9454M & 544.1 & - & 6.1 \\
\quad+\framework & \textbf{0.0022M} & 175.4 & 921.7 & 7.3 \\
\texttt{TimesNet} & 4.6933M & 498.3 & - & 5.9 \\
\quad+\framework & \textbf{0.0022M} & 158.7 & 921.7 & 6.9 \\
\texttt{PatchTST} & 0.9898M & 321.1 & - & 5.4 \\
\quad+\framework & \textbf{0.0022M} & 131.2 & 921.7 & 6.2 \\
\texttt{DLinear} & 0.2265M & 275.3 & - & 1.3 \\
\quad+\framework & \textbf{0.0022M} & 114.9 & 921.7 & 2.1 \\
\bottomrule
\end{tabular}
\end{table}

%% file: tables/training_data_proportion.tex
\begin{table}[t]
\centering
\footnotesize
\caption{Robustness analysis of \framework under temporal distribution shift. MSE results are obtained using PatchTST as the backbone with a missing rate $r=0.5$, averaged across four sequence lengths ${96,192,336,720}$. Best results are \textbf{bolded}.}
\label{tab:training_data_proportion}
\begin{tabular}{l l ccc}
\toprule
\textbf{Dataset} & \textbf{Model} & \textbf{25\%} & \textbf{50\%} & \textbf{100\%} \\
\midrule
\multirow{2}{*}{\textbf{ETTh1}} & \texttt{PatchTST} & 0.204 & 0.188 & 0.165 \\
& \quad+\framework & \textbf{0.173} & \textbf{0.157} & \textbf{0.123} \\
\multirow{2}{*}{\textbf{ETTm2}} & \texttt{PatchTST} & 0.059 & 0.051 & 0.040 \\
& \quad+\framework & \textbf{0.055} & \textbf{0.046} & \textbf{0.037} \\
\multirow{2}{*}{\textbf{Weather}} & \texttt{PatchTST} & \textbf{0.134} & 0.121 & 0.109 \\
& \quad+\framework & \textbf{0.134} & \textbf{0.120} & \textbf{0.107} \\
\multirow{2}{*}{\textbf{Electricity}} & \texttt{PatchTST} & 0.071 & 0.062 & 0.054 \\
& \quad+\framework & \textbf{0.060} & \textbf{0.056} & \textbf{0.039} \\
\bottomrule
\end{tabular}
\end{table}

%% file: tables/infrequent_patterns.tex
\begin{table}[t]
\centering
\caption{Imputation performance in terms of MSE under varying degrees of pattern rarity in the training set. Results are averaged across 360 independent evaluation segments. Best results are \textbf{bolded}.}
\label{tab:infrequent_patterns}
\footnotesize
\setlength{\tabcolsep}{10pt}
\renewcommand{\arraystretch}{1.2}
\begin{tabular}{@{}cccc@{}}
\toprule
\textbf{Pattern occurrences} & 1 & 2 & 4 \\
\midrule
\texttt{DLinear}    & 0.191 & 0.180 & 0.168 \\
\quad+\framework & \textbf{0.164} & \textbf{0.163} & \textbf{0.159} \\
\texttt{TimesNet}   & 0.127 & 0.109 & 0.104 \\
\quad+\framework & \textbf{0.116} & \textbf{0.103} & \textbf{0.099} \\
\texttt{PatchTST}   & 0.141 & 0.133 & 0.128 \\
\quad+\framework & \textbf{0.120} & \textbf{0.118} & \textbf{0.115} \\
\bottomrule
\end{tabular}
\end{table}

%% file: tables/full_results_mse.tex
\begin{table*}
\centering\

\renewcommand{\arraystretch}{1.6}
\setlength{\tabcolsep}{2.5pt}
\caption{{Detailed MSE results  for ten baselines with and without \framework. Lower metric values indicate better performance and best
performances are \textbf{bolded}. The missing rates are set as $r \in \{0.125, 0.25, 0.375, 0.5\}$. The final results are averaged across four different sequence lengths $L \in \{96, 192, 336, 720\}$. \texttt{*} represents variants with the ``former'' suffix.}}
\label{tab:full_results_mse}
\vskip 0.1in
\resizebox{\textwidth}{!}{
\begin{tabular}{c>{\centering\arraybackslash}p{1.0cm}cccccc|cccccccccccccc}

\cline{1-22}

\multicolumn{2}{c}{Baseline type} &

  \multicolumn{6}{c|}{\textbf{Imputation-specialized}} &

  \multicolumn{14}{c}{\textbf{General-purpose}} \\ \cline{3-22}
 Dataset &
  r &
  \texttt{Helix} &
+\framework &
\texttt{SAITS} &
+\framework &
\texttt{Glocal-Ib} &
+\framework &
\texttt{ModernTCN} &
+\framework &
\texttt{TimesNet} &
+\framework &
\texttt{Cross*} &
+\framework &
\texttt{Auto*} &
+\framework &
\texttt{PatchTST} &
+\framework &
\texttt{DLinear} &
+\framework &
\texttt{RLinear} &
+\framework \\ \hline
 &
  0.125 &
  0.117 &
  \textbf{0.105} &
  0.226 &
  \textbf{0.181} &
  0.212 &
  \textbf{0.184} &
  0.037 &
  \textbf{0.034} &
  0.069 &
  \textbf{0.064} &
  0.116 &
  \textbf{0.059} &
  0.065 &
  \textbf{0.049} &
  0.117 &
  \textbf{0.075} &
  0.106 &
  \textbf{0.071} &
  0.107 &
  \textbf{0.078} \\
 &
  0.25 &
  0.137 &
  \textbf{0.137} &
  0.290 &
  \textbf{0.267} &
  0.277 &
  \textbf{0.253} &
  0.046 &
  \textbf{0.041} &
  0.089 &
  \textbf{0.081} &
  0.132 &
  \textbf{0.077} &
  0.083 &
  \textbf{0.060} &
  0.130 &
  \textbf{0.086} &
  0.140 &
  \textbf{0.093} &
  0.139 &
  \textbf{0.101} \\
 &
  0.375 &
  0.148 &
  \textbf{0.141} &
  0.332 &
  \textbf{0.306} &
  0.342 &
  \textbf{0.305} &
  0.058 &
  \textbf{0.055} &
  0.106 &
  \textbf{0.095} &
  0.150 &
  \textbf{0.093} &
  0.106 &
  \textbf{0.080} &
  0.142 &
  \textbf{0.103} &
  0.174 &
  \textbf{0.120} &
  0.173 &
  \textbf{0.131} \\
\multirow{-4}{*}{ETTh1} &
  0.5 &
  0.179 &
  \textbf{0.169} &
  0.437 &
  \textbf{0.383} &
  0.423 &
  \textbf{0.376} &
  0.073 &
  \textbf{0.067} &
  0.128 &
  \textbf{0.112} &
  0.170 &
  \textbf{0.121} &
  0.129 &
  \textbf{0.101} &
  0.165 &
  \textbf{0.123} &
  0.210 &
  \textbf{0.151} &
  0.208 &
  \textbf{0.163} \\ \hline
 &
  0.125 &
  0.109 &
  \textbf{0.103} &
  0.251 &
  \textbf{0.227} &
  0.145 &
  \textbf{0.122} &
  0.040 &
  \textbf{0.039} &
  0.051 &
  \textbf{0.048} &
  0.122 &
  \textbf{0.086} &
  0.048 &
  \textbf{0.044} &
  0.064 &
  \textbf{0.062} &
  0.106 &
  \textbf{0.080} &
  0.104 &
  \textbf{0.088} \\
 &
  0.25 &
  0.082 &
  \textbf{0.080} &
  0.259 &
  \textbf{0.225} &
  0.316 &
  \textbf{0.283} &
  0.044 &
  \textbf{0.043} &
  0.063 &
  \textbf{0.059} &
  0.139 &
  \textbf{0.102} &
  0.054 &
  \textbf{0.049} &
  0.069 &
  \textbf{0.068} &
  0.138 &
  \textbf{0.102} &
  0.133 &
  \textbf{0.101} \\
 &
  0.375 &
  0.094 &
  \textbf{0.091} &
  0.296 &
  \textbf{0.271} &
  0.260 &
  \textbf{0.231} &
  0.048 &
  \textbf{0.047} &
  0.071 &
  \textbf{0.066} &
  0.158 &
  \textbf{0.124} &
  0.061 &
  \textbf{0.057} &
  0.074 &
  \textbf{0.071} &
  0.169 &
  \textbf{0.123} &
  0.159 &
  \textbf{0.119} \\
\multirow{-4}{*}{ETTh2} &
  0.5 &
  0.123 &
  \textbf{0.121} &
  0.323 &
  \textbf{0.284} &
  0.407 &
  \textbf{0.379} &
  0.057 &
  \textbf{0.052} &
  0.083 &
  \textbf{0.076} &
  0.183 &
  \textbf{0.151} &
  0.074 &
  \textbf{0.071} &
  0.080 &
  \textbf{0.079} &
  0.200 &
  \textbf{0.151} &
  0.185 &
  \textbf{0.141} \\ \hline
 &
  0.125 &
  0.043 &
  \textbf{0.040} &
  0.081 &
  \textbf{0.062} &
  0.120 &
  \textbf{0.109} &
  0.019 &
  \textbf{0.017} &
  0.025 &
  \textbf{0.024} &
  0.052 &
  \textbf{0.041} &
  0.054 &
  \textbf{0.030} &
  0.067 &
  \textbf{0.039} &
  0.053 &
  \textbf{0.030} &
  0.053 &
  \textbf{0.030} \\
 &
  0.25 &
  0.050 &
  \textbf{0.048} &
  0.116 &
  \textbf{0.098} &
  0.161 &
  \textbf{0.150} &
  0.024 &
  \textbf{0.021} &
  0.034 &
  \textbf{0.033} &
  0.054 &
  \textbf{0.043} &
  0.037 &
  \textbf{0.022} &
  0.066 &
  \textbf{0.038} &
  0.071 &
  \textbf{0.043} &
  0.070 &
  \textbf{0.042} \\
 &
  0.375 &
  0.057 &
  \textbf{0.053} &
  0.147 &
  \textbf{0.112} &
  0.208 &
  \textbf{0.191} &
  0.026 &
  \textbf{0.022} &
  0.036 &
  \textbf{0.035} &
  0.059 &
  \textbf{0.051} &
  0.079 &
  \textbf{0.040} &
  0.073 &
  \textbf{0.041} &
  0.091 &
  \textbf{0.060} &
  0.089 &
  \textbf{0.058} \\
\multirow{-4}{*}{ETTm1} &
  0.5 &
  0.058 &
  \textbf{0.051} &
  0.190 &
  \textbf{0.154} &
  0.262 &
  \textbf{0.248} &
  0.028 &
  \textbf{0.026} &
  0.041 &
  \textbf{0.039} &
  0.064 &
  \textbf{0.054} &
  0.088 &
  \textbf{0.047} &
  0.083 &
  \textbf{0.054} &
  {0.117} &
  {\textbf{0.093}} &
  {0.106} &
  {\textbf{0.059}} \\ \hline
 &
  0.125 &
  0.021 &
  \textbf{0.018} &
  0.082 &
  \textbf{0.065} &
  0.052 &
  \textbf{0.043} &
  0.021 &
  0.021 &
  0.020 &
  0.020 &
  0.067 &
  \textbf{0.056} &
  0.061 &
  \textbf{0.046} &
  0.032 &
  \textbf{0.030} &
  0.064 &
  \textbf{0.050} &
  0.061 &
  \textbf{0.049} \\
 &
  0.25 &
  0.021 &
  \textbf{0.021} &
  0.104 &
  \textbf{0.088} &
  0.075 &
  \textbf{0.063} &
  0.021 &
  \textbf{0.020} &
  0.022 &
  0.022 &
  0.080 &
  \textbf{0.066} &
  0.065 &
  \textbf{0.049} &
  0.035 &
  \textbf{0.033} &
  0.084 &
  \textbf{0.065} &
  0.079 &
  \textbf{0.063} \\
 &
  0.375 &
  0.019 &
  \textbf{0.017} &
  0.136 &
  \textbf{0.115} &
  0.120 &
  \textbf{0.112} &
  0.025 &
  \textbf{0.024} &
  0.024 &
  \textbf{0.023} &
  0.086 &
  \textbf{0.074} &
  0.045 &
  \textbf{0.041} &
  0.037 &
  \textbf{0.036} &
  0.105 &
  \textbf{0.091} &
  0.095 &
  \textbf{0.078} \\
\multirow{-4}{*}{ETTm2} &
  0.5 &
  0.023 &
  \textbf{0.021} &
  0.113 &
  \textbf{0.101} &
  0.327 &
  \textbf{0.305} &
  \textbf{0.026} &
  0.027 &
  0.027 &
  \textbf{0.026} &
  0.100 &
  \textbf{0.088} &
  0.045 &
  \textbf{0.042} &
  0.040 &
  \textbf{0.037} &
  0.127 &
  \textbf{0.101} &
  0.112 &
  \textbf{0.089} \\ \hline
 &
  0.125 &
  0.102 &
  \textbf{0.089} &
  0.204 &
  \textbf{0.189} &
  0.313 &
  \textbf{0.288} &
  0.056 &
  \textbf{0.052} &
  0.098 &
  \textbf{0.088} &
  0.068 &
  \textbf{0.061} &
  0.138 &
  \textbf{0.129} &
  0.091 &
  \textbf{0.081} &
  0.063 &
  \textbf{0.052} &
  0.063 &
  \textbf{0.051} \\
 &
  0.25 &
  0.108 &
  \textbf{0.104} &
  0.204 &
  \textbf{0.187} &
  0.320 &
  \textbf{0.301} &
  0.072 &
  \textbf{0.066} &
  0.095 &
  \textbf{0.086} &
  0.101 &
  \textbf{0.089} &
  0.138 &
  \textbf{0.127} &
  0.088 &
  \textbf{0.083} &
  0.103 &
  \textbf{0.083} &
  0.094 &
  \textbf{0.077} \\
 &
  0.375 &
  0.121 &
  \textbf{0.117} &
  0.208 &
  \textbf{0.185} &
  0.329 &
  \textbf{0.310} &
  0.081 &
  \textbf{0.078} &
  0.097 &
  \textbf{0.088} &
  0.115 &
  \textbf{0.102} &
  0.145 &
  \textbf{0.133} &
  0.094 &
  \textbf{0.089} &
  0.128 &
  \textbf{0.106} &
  0.115 &
  \textbf{0.095} \\
\multirow{-4}{*}{Electricity} &
  0.5 &
  0.126 &
  \textbf{0.122} &
  0.220 &
  \textbf{0.193} &
  0.338 &
  \textbf{0.329} &
  0.089 &
  \textbf{0.083} &
  0.100 &
  \textbf{0.090} &
  0.129 &
  \textbf{0.108} &
  0.151 &
  \textbf{0.138} &
  0.109 &
  \textbf{0.107} &
  0.155 &
  \textbf{0.127} &
  0.138 &
  \textbf{0.116} \\ \hline
 &
  0.125 &
  0.027 &
  \textbf{0.023} &
  0.029 &
  \textbf{0.027} &
  0.080 &
  \textbf{0.069} &
  0.029 &
  \textbf{0.026} &
  0.027 &
  \textbf{0.026} &
  0.038 &
  \textbf{0.034} &
  0.032 &
  \textbf{0.026} &
  0.034 &
  \textbf{0.029} &
  0.038 &
  \textbf{0.032} &
  0.038 &
  \textbf{0.032} \\
 &
  0.25 &
  0.031 &
  \textbf{0.029} &
  0.029 &
  \textbf{0.026} &
  0.091 &
  \textbf{0.085} &
  0.030 &
  \textbf{0.029} &
  0.032 &
  \textbf{0.031} &
  0.038 &
  \textbf{0.036} &
  0.037 &
  \textbf{0.031} &
  0.046 &
  \textbf{0.035} &
  0.047 &
  \textbf{0.038} &
  0.047 &
  \textbf{0.038} \\
 &
  0.375 &
  0.034 &
  \textbf{0.030} &
  0.072 &
  \textbf{0.064} &
  0.133 &
  \textbf{0.124} &
  0.033 &
  \textbf{0.032} &
  0.036 &
  \textbf{0.035} &
  0.042 &
  \textbf{0.039} &
  0.066 &
  \textbf{0.058} &
  0.039 &
  \textbf{0.033} &
  0.055 &
  \textbf{0.042} &
  0.054 &
  \textbf{0.042} \\
\multirow{-4}{*}{Weather} &
  0.5 &
  0.038 &
  \textbf{0.034} &
  0.036 &
  \textbf{0.033} &
  0.179 &
  \textbf{0.166} &
  {0.036} &
  0.036 &
  0.038 &
  \textbf{0.037} &
  0.044 &
  \textbf{0.040} &
  0.050 &
  \textbf{0.042} &
  0.054 &
  \textbf{0.039} &
  0.064 &
  \textbf{0.050} &
  0.062 &
  \textbf{0.048} \\ \hline
\end{tabular}
}
\end{table*}

%% file: tables/full_results_mae.tex
\begin{table*}
\centering\

\renewcommand{\arraystretch}{1.6}
\setlength{\tabcolsep}{2.5pt}
\caption{{Detailed MAE results for ten baselines with and without \framework. Lower metric values indicate better performance and best
performances are \textbf{bolded}. The missing rates are set as $r \in \{0.125, 0.25, 0.375, 0.5\}$. The final results are averaged across four different sequence lengths $L \in \{96, 192, 336, 720\}$. \texttt{*} represents variants with the ``former'' suffix.}}
\label{tab:full_results_mae}
\vskip 0.1in
\resizebox{\textwidth}{!}{
\begin{tabular}{c>{\centering\arraybackslash}p{1.0cm}cccccc|cccccccccccccc}

\cline{1-22}

\multicolumn{2}{c}{Baseline type} &
  \multicolumn{6}{c|}{\textbf{Imputation-specialized}} &
  \multicolumn{14}{c}{\textbf{General-purpose}} \\ \cline{3-22}
  Dataset & r &
  \texttt{Helix} & +\framework &
  \texttt{SAITS} & +\framework &
  \texttt{Glocal-Ib} & +\framework &
  \texttt{ModernTCN} & +\framework &
  \texttt{TimesNet} & +\framework &
  \texttt{Cross*} & +\framework &
  \texttt{Auto*} & +\framework &
  \texttt{PatchTST} & +\framework &
  \texttt{DLinear} & +\framework &
  \texttt{RLinear} & +\framework \\ \hline

 & 0.125 &
  0.220 & \textbf{0.209} & 0.311 & \textbf{0.249} & 0.302 & \textbf{0.261} &
  0.134 & \textbf{0.127} & 0.177 & \textbf{0.171} &
  0.236 & \textbf{0.170} & 0.176 & \textbf{0.146} &
  0.224 & \textbf{0.182} & 0.227 & \textbf{0.184} & 0.228 & \textbf{0.193} \\
 & 0.25 &
  0.241 & \textbf{0.238} & 0.348 & \textbf{0.301} & 0.347 & \textbf{0.311} &
  0.149 & \textbf{0.133} & 0.202 & \textbf{0.196} &
  0.253 & \textbf{0.194} & 0.198 & \textbf{0.162} &
  0.236 & \textbf{0.196} & 0.261 & \textbf{0.211} & 0.261 & \textbf{0.217} \\
 & 0.375 &
  0.251 & \textbf{0.242} & 0.383 & \textbf{0.342} & 0.389 & \textbf{0.347} &
  0.166 & \textbf{0.159} & 0.221 & \textbf{0.215} &
  0.271 & \textbf{0.213} & 0.225 & \textbf{0.188} &
  0.249 & \textbf{0.214} & 0.290 & \textbf{0.239} & 0.289 & \textbf{0.245} \\
\multirow{-4}{*}{ETTh1} & 0.5 &
  0.273 & \textbf{0.256} & 0.436 & \textbf{0.361} & 0.433 & \textbf{0.380} &
  0.184 & \textbf{0.174} & 0.241 & \textbf{0.237} &
  0.291 & \textbf{0.243} & 0.246 & \textbf{0.211} &
  0.265 & \textbf{0.233} & 0.318 & \textbf{0.267} & 0.316 & \textbf{0.273} \\ \hline

 & 0.125 &
  0.234 & \textbf{0.226} & 0.343 & \textbf{0.288} & 0.262 & \textbf{0.229} &
  0.131 & \textbf{0.127} & 0.152 & \textbf{0.145} &
  0.231 & \textbf{0.202} & \textbf{0.147} & 0.148 &
  0.167 & \textbf{0.165} & 0.220 & \textbf{0.197} & 0.218 & 0.218 \\
 & 0.25 &
  0.202 & \textbf{0.195} & 0.350 & \textbf{0.313} & 0.433 & \textbf{0.387} &
  0.138 & \textbf{0.135} & 0.171 & \textbf{0.159} &
  0.249 & \textbf{0.219} & \textbf{0.156} & 0.157 &
  0.172 & \textbf{0.170} & 0.253 & \textbf{0.222} & 0.249 & \textbf{0.223} \\
 & 0.375 &
  0.220 & \textbf{0.214} & 0.372 & \textbf{0.329} & 0.370 & \textbf{0.335} &
  0.143 & \textbf{0.142} & 0.182 & \textbf{0.169} &
  0.267 & \textbf{0.240} & \textbf{0.165} & 0.167 &
  0.179 & \textbf{0.174} & 0.280 & \textbf{0.244} & 0.273 & \textbf{0.242} \\
\multirow{-4}{*}{ETTh2} & 0.5 &
  0.254 & \textbf{0.253} & 0.396 & \textbf{0.338} & 0.478 & \textbf{0.428} &
  0.159 & \textbf{0.152} & 0.196 & \textbf{0.185} &
  0.290 & \textbf{0.265} & \textbf{0.183} & 0.184 &
  \textbf{0.186} & 0.187 & 0.306 & \textbf{0.269} & 0.296 & \textbf{0.263} \\ \hline

 & 0.125 &
  0.124 & \textbf{0.117} & 0.181 & \textbf{0.150} & 0.220 & \textbf{0.201} &
  0.093 & \textbf{0.088} & 0.099 & \textbf{0.095} &
  0.160 & \textbf{0.148} & 0.158 & \textbf{0.114} &
  0.170 & \textbf{0.114} & 0.158 & \textbf{0.121} & 0.157 & \textbf{0.122} \\
 & 0.25 &
  0.136 & \textbf{0.131} & 0.209 & \textbf{0.167} & 0.250 & \textbf{0.236} &
  0.103 & \textbf{0.097} & 0.111 & 0.111 &
  0.163 & \textbf{0.128} & 0.131 & \textbf{0.115} &
  0.171 & \textbf{0.126} & 0.184 & \textbf{0.145} & 0.183 & \textbf{0.144} \\
 & 0.375 &
  0.146 & \textbf{0.142} & 0.233 & \textbf{0.202} & 0.281 & \textbf{0.264} &
  0.106 & \textbf{0.098} & 0.122 & \textbf{0.119} &
  0.171 & \textbf{0.164} & 0.189 & \textbf{0.132} &
  0.180 & \textbf{0.137} & 0.209 & \textbf{0.171} & 0.207 & \textbf{0.167} \\
\multirow{-4}{*}{ETTm1} & 0.5 &
  0.147 & \textbf{0.140} & 0.267 & \textbf{0.211} & 0.314 & \textbf{0.287} &
  0.118 & \textbf{0.110} & 0.145 & \textbf{0.136} &
  0.192 & \textbf{0.176} & 0.196 & \textbf{0.154} &
  0.190 & \textbf{0.150} & 0.226 & \textbf{0.205} & 0.215 & \textbf{0.179} \\ \hline

 & 0.125 &
  0.102 & \textbf{0.096} & 0.199 & \textbf{0.176} & 0.164 & \textbf{0.141} &
  0.088 & \textbf{0.087} & 0.085 & \textbf{0.084} &
  0.174 & \textbf{0.163} & 0.157 & \textbf{0.143} &
  \textbf{0.112} & 0.125 & 0.168 & \textbf{0.150} & 0.165 & \textbf{0.149} \\
 & 0.25 &
  0.095 & \textbf{0.094} & 0.217 & \textbf{0.189} & 0.198 & \textbf{0.175} &
  0.089 & \textbf{0.087} & 0.091 & \textbf{0.090} &
  0.187 & \textbf{0.173} & 0.154 & \textbf{0.141} &
  0.117 & \textbf{0.113} & 0.195 & \textbf{0.172} & 0.189 & \textbf{0.171} \\
 & 0.375 &
  0.094 & \textbf{0.091} & 0.244 & \textbf{0.209} & 0.250 & \textbf{0.237} &
  0.098 & \textbf{0.097} & 0.096 & \textbf{0.095} &
  0.192 & \textbf{0.181} & 0.136 & \textbf{0.132} &
  0.122 & \textbf{0.119} & 0.218 & \textbf{0.197} & 0.209 & \textbf{0.186} \\
\multirow{-4}{*}{ETTm2} & 0.5 &
  0.103 & \textbf{0.098} & 0.229 & \textbf{0.207} & 0.445 & \textbf{0.403} &
  \textbf{0.100} & 0.102 & \textbf{0.104} & 0.106 &
  0.206 & \textbf{0.194} & 0.139 & \textbf{0.134} &
  0.128 & \textbf{0.124} & 0.241 & \textbf{0.215} & 0.227 & \textbf{0.205} \\ \hline

 & 0.125 &
  0.191 & \textbf{0.177} & 0.309 & \textbf{0.278} & 0.381 & \textbf{0.354} &
  0.166 & \textbf{0.159} & 0.217 & \textbf{0.203} &
  0.184 & \textbf{0.169} & 0.258 & \textbf{0.216} &
  0.222 & \textbf{0.214} & 0.176 & \textbf{0.151} & 0.176 & \textbf{0.148} \\
 & 0.25 &
  0.197 & \textbf{0.192} & 0.305 & \textbf{0.279} & 0.387 & \textbf{0.353} &
  0.191 & \textbf{0.181} & 0.211 & \textbf{0.207} &
  0.228 & \textbf{0.209} & 0.256 & \textbf{0.221} &
  0.219 & \textbf{0.202} & 0.231 & \textbf{0.202} & 0.219 & \textbf{0.187} \\
 & 0.375 &
  0.211 & \textbf{0.206} & 0.307 & \textbf{0.278} & 0.394 & \textbf{0.366} &
  0.202 & \textbf{0.200} & 0.212 & \textbf{0.206} &
  0.244 & \textbf{0.215} & 0.263 & \textbf{0.227} &
  0.227 & \textbf{0.220} & 0.258 & \textbf{0.215} & 0.244 & \textbf{0.212} \\
\multirow{-4}{*}{Electricity} & 0.5 &
  0.218 & \textbf{0.211} & 0.314 & \textbf{0.281} & 0.401 & \textbf{0.381} &
  0.211 & \textbf{0.204} & 0.218 & \textbf{0.213} &
  0.260 & \textbf{0.221} & 0.267 & \textbf{0.232} &
  0.245 & \textbf{0.241} & 0.286 & \textbf{0.241} & 0.269 & \textbf{0.238} \\ \hline

 & 0.125 &
  0.048 & \textbf{0.041} & 0.039 & \textbf{0.035} & 0.115 & \textbf{0.108} &
  0.054 & \textbf{0.052} & 0.056 & \textbf{0.055} &
  0.093 & \textbf{0.079} & 0.067 & \textbf{0.058} &
  0.064 & \textbf{0.056} & 0.088 & \textbf{0.071} & 0.089 & \textbf{0.072} \\
 & 0.25 &
  0.052 & \textbf{0.049} & 0.040 & \textbf{0.039} & 0.132 & \textbf{0.125} &
  0.058 & \textbf{0.056} & 0.065 & \textbf{0.063} &
  0.095 & \textbf{0.088} & 0.078 & \textbf{0.069} &
  0.086 & \textbf{0.060} & 0.107 & \textbf{0.084} & 0.106 & \textbf{0.084} \\
 & 0.375 &
  0.062 & \textbf{0.056} & 0.047 & \textbf{0.041} & 0.181 & \textbf{0.167} &
  0.063 & \textbf{0.061} & \textbf{0.071} & 0.072 &
  0.104 & \textbf{0.095} & 0.118 & \textbf{0.106} &
  0.071 & \textbf{0.064} & 0.120 & \textbf{0.091} & 0.118 & \textbf{0.091} \\
\multirow{-4}{*}{Weather} & 0.5 &
  0.067 & \textbf{0.062} & 0.049 & \textbf{0.045} & 0.238 & \textbf{0.221} &
  0.068 & \textbf{0.067} & 0.075 & \textbf{0.074} &
  0.105 & \textbf{0.099} & 0.099 & \textbf{0.087} &
  0.095 & \textbf{0.070} & 0.133 & \textbf{0.103} & 0.129 & \textbf{0.100} \\ \hline

\end{tabular}
}
\end{table*}

%% file: revised_sections/conlusion.tex
\section{Conclusion}
\label{sec:conclusion}

We proposed \framework, a retrieval-augmented framework for time series imputation. \framework addresses the limitation of relying only on corrupted local context by retrieving relevant historical patterns to support missing-value reconstruction. Its core component, Latent Embedding Alignment (LEA), mitigates the representation mismatch between corrupted queries and clean historical candidates while maintaining efficient cached retrieval. Through a lightweight model-agnostic adapter, \framework can be integrated with different imputation backbones. Experiments on multiple real-world benchmarks show that \framework consistently improves strong baseline models across different sequence lengths and missing rates.
In future work, we plan to explore more efficient retrieval strategies for large-scale historical databases. We also aim to extend \framework to more challenging missingness patterns, such as block missingness and sensor-level failures.